# Machine Learning Based Analytics for the Significance of Gait Analysis in Monitoring and Managing Lower Extremity Injuries

**Mostafa Rezapour, PhD[1*]; Rachel B. Seymour, PhD[2]; Stephen H. Sims, MD[2]; Madhav A. Karunakar, MD [2]; Nahir Habet, MS[2]; Metin Nafi Gurcan, PhD[1]**

[1]Center for Artificial Intelligence, Wake Forest School of Medicine, Winston-Salem, NC, USA

[2] Department of Orthopaedic Surgery, Atrium Health Musculoskeletal Institute and Wake Forest University School of Medicine, Charlotte, NC, USA

**\* Correspondence:**
Mostafa Rezapour, PhD
mrezapou@wakehealth.edu

## Abstract

**Objective:** This study aimed to explore the potential of gait analysis as a predictive tool for assessing post-injury complications, e.g., infection, malunion, or hardware irritation, in patients with lower extremity fractures. More precisely, the research focused on determining the proficiency of supervised machine learning models in predicting complications using two consecutive gait datasets.

**Methods:** We prospectively identified patients with lower extremity fractures at a tertiary academic center. These patients underwent gait analysis with a chest-mounted IMU device. Using customized software, the raw gait data was preprocessed, emphasizing 12 essential gait variables. The data were standardized, and several machine learning models including XGBoost, Logistic Regression, SVM, LightGBM, and Random Forest were trained, tested, and evaluated. Special attention was given to class imbalance, addressed using SMOTE. Additionally, we introduced a novel methodology to compute the Rate of Change (ROC) for gait variables, which operates independently of the time difference between the gait analyses of different patients.

**Results:** XGBoost was identified as the optimal model both before and after the application of SMOTE. Prior to using SMOTE, the model achieved an average test AUC of 0.90 (95% CI: [0.79, 1.00]) and an average test accuracy of 86% (95% CI: [75%, 97%]). Through feature importance analysis, a pivotal role was attributed to the duration between the occurrence of the injury and the initial gait analysis. Data patterns over time revealed early aggressive physiological compensations, followed by stabilization phases, underscoring the importance of prompt gait analysis.

**Conclusion:** This study emphasizes the transformative potential of using machine learning, particularly XGBoost, in gait analysis for orthopedic care. By effectively predicting post-injury complications, early gait assessment becomes pivotal, revealing timely intervention points. The findings underscore a paradigm shift in orthopedics towards a more data-informed, proactive approach, paving the way for enhanced patient outcomes.

## 1    Introduction

Orthopedic injuries to the lower extremity are commonplace in both military and civilian contexts. The ultimate aim of surgical and rehabilitative teams is the patient's return to their prior functional level, which can be a challenging balance given multiple short-term goals and associated risks [1]. Rehabilitation, especially for the lower extremities, is an intricate process. It focuses on early movement, improving range of motion, and careful weight-bearing, but must be undertaken with caution to prevent harm to fracture sites and soft tissues. In the rehabilitation journey, patient compliance, pain management, and addressing mobility challenges are all pivotal elements. Especially in older demographics, reduced mobility can precipitate a host of health concerns ranging from physical deterioration to thromboembolic events and respiratory complications [1].

Historically, clinicians were reserved in recommending early weight-bearing for patients with lower extremity fractures, wary of complications such as implant failure [2]. However, contemporary trends suggest a paradigm shift, leaning towards early mobilization in certain cases. Tailoring rehab strategies, accounting for factors like patient compliance, pain, and immobility, is essential for the best patient outcomes [2].

Different fractures necessitate distinct weight-bearing guidelines. Calcaneal fractures: ORIF often demands a 6–12-week NWB period, whereas external fixation permits immediate weight-bearing [3, 4]. Tibial plafond and shaft fractures: Guidelines vary based on the method of fixation, the fracture's nature, and severity [5, 6, 7]. Femoral fractures: ORIF and certain femoral shaft fractures typically follow a 6–12-week NWB protocol, but some methods allow for immediate weight-bearing [8, 9, 10]. Pelvic fractures: The weight-bearing protocol depends on the fracture's stability and the chosen fixation approach [11, 12, 13].

Weight-bearing guidelines blend clinical expertise and research evidence, considering the injury specifics, fixation techniques, and recommended non-weight-bearing durations. But, there's a scarcity of data on recovery rates and returning to pre-injury functionality post such fractures. This is where gait analysis, a tool assessing walking patterns, becomes invaluable, offering insights into a patient's rehabilitation journey [14].

The evolution of clinical gait analysis started in the 1980s with the pioneering labs by the United Technologies Corporation and Oxford Dynamics [15]. These labs introduced retro-reflective markers detected by modified cameras, leading to the development of three-dimensional positional tracking of these markers. The technology has advanced significantly, but interpretation models, such as the Conventional Gait Model, continue to evolve [16]. Traditional floor-based photocell systems and multi-camera setups have been standard for gait assessment [17], yet wearable inertial measurement units (IMUs) offer a cost-effective and reliable alternative in clinical settings [18].

In our previous works, we utilized machine learning to investigate the challenges posed by Covid-19 [19, 20, 21] and to diagnose cancer from Whole Slide Images [22, 23, 24]. In this study, we introduce machine learning models to predict orthopedic complications, including infection, malunion, and hardware irritation. To the best of our knowledge, this is the first time such predictions have been made.

## 2    Methods

Our research secured approval from three consecutive Institutional Review Boards (IRBs): Carolinas HealthCare System IRB on March 18, 2015, (IRB File #03-15-11E); Atrium Health IRB; and currently, the Wake Forest University School of Medicine IRB (IRB00082570). We have consistently maintained this approval, adapting to institutional name changes. All research methods adhered to the guidelines and regulations of these institutions. Although our study, titled "Mobility Toolkit Development in the Orthopedic Surgery Patient Population," didn't involve experimental interventions, all its protocols, including data collection, were vetted by the IRBs. Written informed consent was obtained from all participants or their legal guardians before any study-related activities. Patients were initially enrolled prior to performance testing, followed at standard care visits by the treating surgeon, and subsequently invited for the study during post-operative clinic visits. Measures to ensure data confidentiality included secure storage and restricted access, with strict adherence to ethical considerations for data protection and participant anonymity throughout the research.

The study included a diverse group of participants: adult patients aged 19 years and older. Some patients experience major orthopedic complications such as infection, non-union/malunion, and hardware irritation. After removing outliers, **Figure 1** illustrates the demographic distribution of patients selected for this study.

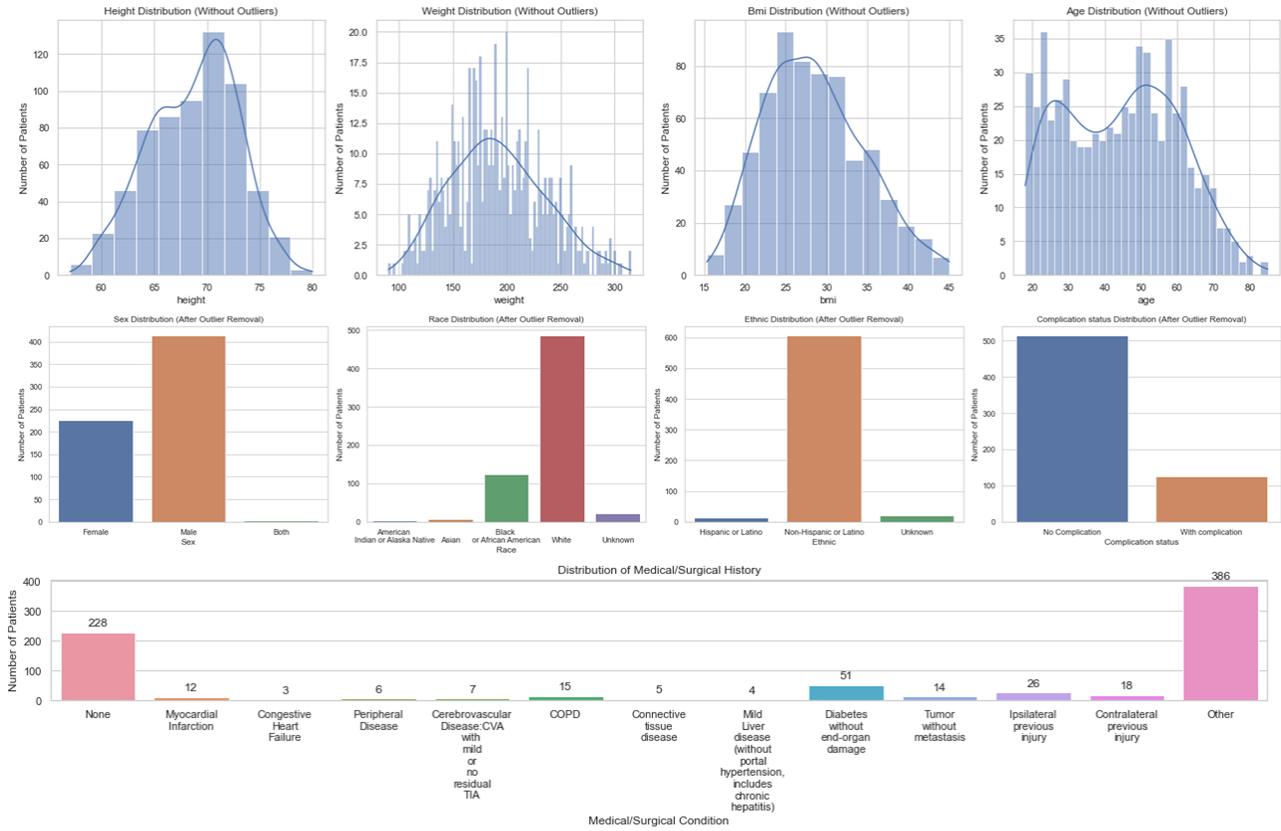

**Figure 1.** Demographic distribution of study participants after outlier removal.

Throughout the course of patients' recovery, gait analysis was performed using a well-validated chest mounted IMU device, and the patients were also requested to complete self-reported outcome measures [18]. To effectively process and examine the collected data, we developed customized software that underwent preprocessing of the raw data. Subsequently, we focused our analysis on 12 distinct gait variables, each of which carries significant implications for movement efficiency, coordination, power generation, gait symmetry, stability, balance, temporal dynamics, and body orientation. These variables enable us to gain a comprehensive understanding of gait dynamics and body movements during the act of walking. These gait variables and their descriptions are summarized in **Table 1.**

**Table 1.** Summary of gait variables and descriptions.

| **Variable** | **Description** |
|---|---|
| Mean Left Leg Lift Acceleration | The average acceleration of the left leg during the lifting phase of the gait cycle. |
| Left Leg Lift Acceleration SEM | Standard Error Mean (SEM) of the acceleration of the left leg during the lifting phase of the gait cycle. |
| Mean Right Leg Lift Acceleration | The average acceleration of the right leg during the lifting phase of the gait cycle. |
| Right Leg Lift Acceleration SEM | Standard Error Mean (SEM) of the acceleration of the right leg during the lifting phase of the gait cycle. |
| Mean Left Stance Time | The average duration of the left leg's stance phase, which is the period when the foot is in contact with the ground. |
| Left Stance Time SEM | Standard Error Mean (SEM) of the duration of the left leg's stance phase. |

| Mean Right Stance Time | The average duration of the right leg's stance phase, which is the period when the foot is in contact with the ground. |
|---|---|
| Right Stance Time SEM | Standard Error Mean (SEM) of the duration of the right leg's stance phase. |
| Mean Pitch Magnitude | The average magnitude of the pitch (forward-backward) movement of the body during walking. |
| Pitch Magnitude SEM | Standard Error Mean (SEM) of the magnitude of the pitch (forward-backward) movement of the body during walking. |
| Mean Roll Magnitude | The average magnitude of the roll (side-to-side) movement of the body during walking. |
| Roll Magnitude SEM | Standard Error Mean (SEM) of the magnitude of the roll (side-to-side) movement of the body during walking. |

In this study, our primary focus centers on the application of gait analysis and machine learning to lower extremity injuries. While our initial objective is to assess the statistical relationship between a patient's readmission or experience of complications and their underlying medical conditions, ensuring that underlying conditions do not directly affect orthopedic complications, our emphasis lies on the following objectives:

- Evaluating the importance of early gait analysis for lower extremity injuries through the use of a supervised machine learning model.
- Determining the potential of supervised machine learning models to predict complications a patient might experience using two consecutive gait data sets derived from the IMU.

To investigate the relationship between main complications and clinical factors like underlying conditions, we will conduct statistical tests, including the Chi-squared test [25], on the entire dataset. We will similarly utilize this test to explore the correlation between readmissions and underlying medical conditions. This approach aids in discerning the reasons for readmissions, allowing us to determine if they are attributed to specific underlying conditions, especially if these readmissions are unrelated to the main complications.

To achieve our second and third research objectives, we focus on a subset of patients who underwent gait analysis twice. Rather than having two separate gait observations at potentially different time points for each patient, we will streamline the data. For each patient, we introduce a set of variables by combining the Rate of Change (ROC) for gait variables with the duration in weeks that elapsed between the injury and their initial gait analysis visit (weeks since injury). This approach provides a consistent and consolidated view of the patient's progress, regardless of the differing time intervals between individual patient observations. We introduce the Rate of Change (ROC) gait variable, calculated using Algorithm 1.

**Algorithm 1. The Rate of Change (ROC) for gait variables**

**Input:**
- m: Number of gait variables
- n: Number of patients
- G: gait variable values for each patient and analysis (dimension: n by m)
- W: Weeks between injury and each gait analysis (dimension: n by 2)

**Step 0:** Initialize an empty array LTG (dimension: n by 2)

**Step 1: for** $i = 1, 2, \ldots, n$ do:

    **for** $j = 1, 2, \ldots, m$ do:

- Compute the rate of change of variable j for patient i:

$$\text{ROC}_{ij} = \frac{G_{ij2} - G_{ij1}}{W_2 - W_1},$$

where $G_{ijt}$ and $W_t$ represent the gait variable j value for patient i at analysis t, and the number of weeks between injury and analysis t, respectively.

- Assign $\text{LTG}_{ij} = (W_1, \text{ROC}_{ij})$.

        **End (for)**
    **End (for)**
**Output:** LTG

___

For each gait variable, Algorithm 1 generates a pair indicating the weeks from injury to the first gait analysis and the rate of change in that gait variable between the first and second visits. This process encapsulates both the duration since injury and the healing trajectory in one data point. After calculating the ROC for all gait variables, we assemble a 13-dimensional vector for each patient that includes the weeks since injury leading up to the first gait analysis and the ROC of all gait variables. These variables serve as predictors in supervised machine learning models to predict a patient's risk of complications, such as infection, malunion, and hardware irritation. Note that the duration between the injury and the first gait analysis is also an input for the model. Once we determine a reliable model, we can explore the importance of this duration in predicting complication outcomes.

We adopt a comprehensive approach to model building that includes data preprocessing, model selection, hyperparameter tuning, and evaluation. This rigorous process aims to develop robust and generalizable models by addressing key concerns such as class imbalance and data leakage. Before training our machine learning models, we standardize the data to a mean of zero and a standard deviation of one, facilitating enhanced and quicker convergence for various optimization algorithms. We utilize supervised machine learning models, including logistic regression (LR) [26, 27], support vector machines (SVM) [28, 29], Random Forest [30], XGBoost [31, 32], and LightGBM [33], across two distinct data splitting scenarios.

In the first scenario, we partition the dataset into training-validation and test sets at a 75:25 ratio. This means 75% of the data supports the training and validation phases, with the remaining 25% reserved for testing. This split is stratified based on the target variable, ensuring that both sets closely reflect the target variable's distribution. Given our imbalanced dataset, we've introduced an alternative approach.

In the second scenario, we start by randomly selecting 12 patients from each complication group (both with and without complications) and designate them as the test set, ensuring no exposure to the model during training. The remainder constitutes the training-validation set. During training, we employ the Synthetic Minority Oversampling Technique (SMOTE) [34] to capture the full range of patient variations with complications.

Hyperparameter tuning is executed on the training-validation data. For each classifier, a specific set of hyperparameters is explored. Grid search identifies the best parameters, guided by the AUC-ROC score. This process is bolstered by a 5-fold Stratified Cross-Validation to fortify robustness and diminish overfitting risks. Finally, the bootstrap resampling technique calculates the AUC-ROC score's confidence intervals.

## 3    Results

The results of the Chi-squared tests reveal no evidence of a dependency between the main complication and underlying conditions. However, there is an indication that readmission is dependent on underlying medical conditions. **Figure 2** displays the p-value, the contingency matrix, and the distribution of patients with underlying conditions who experienced complications or were readmitted, segmented by age groups. **Figure 3** illustrates the distribution of patients based on their fracture types, readmission status, complication status, and underlying conditions, considering only those with two gait analysis visits for the supervised machine learning objectives.

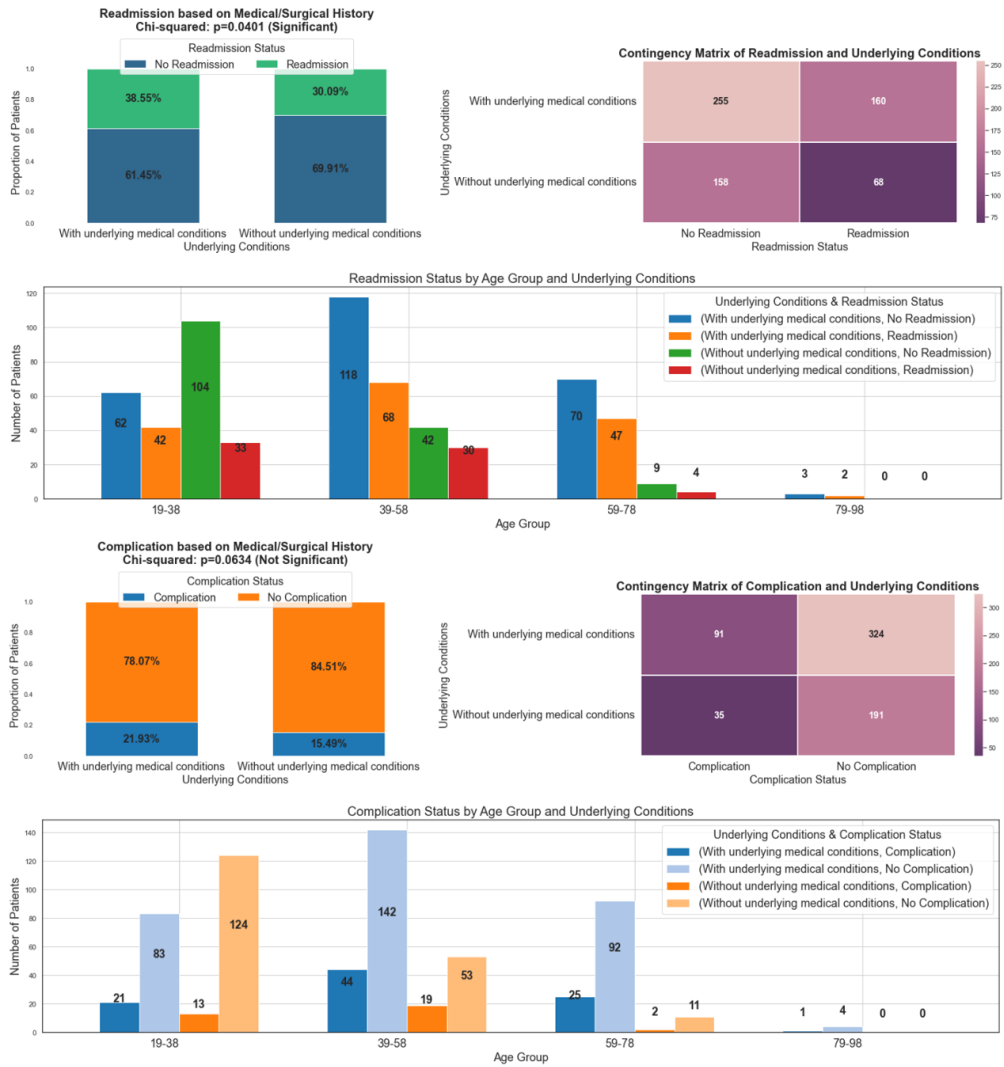

**Figure 2.** Analysis of Dependency between Underlying Conditions, Complications, and Readmission Rates Segmented by Age Groups. The visualization includes a contingency matrix, p-value assessments, and a distribution plot highlighting the correlation of complications or readmission with underlying medical conditions.

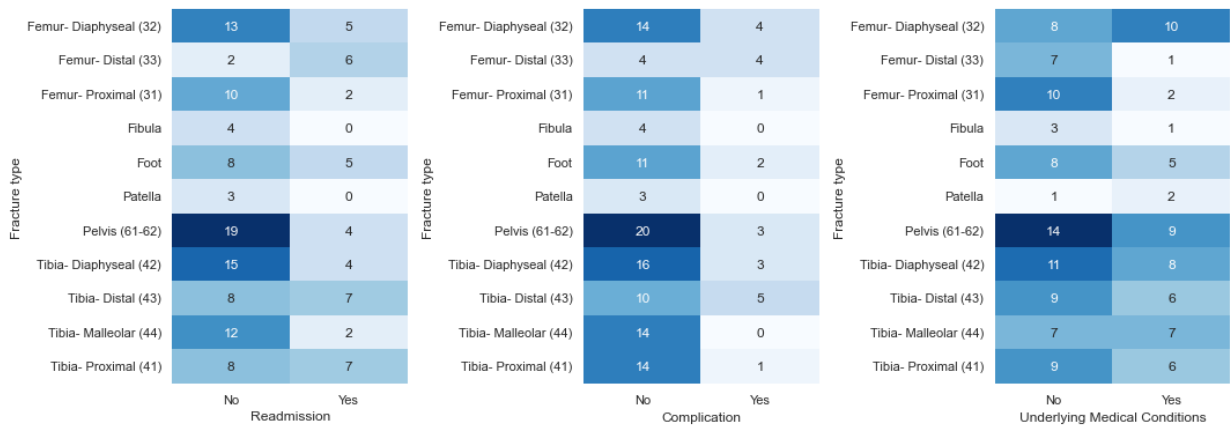

**Figure 3.** Distribution of patients by fracture type, readmission status, complication status, and underlying conditions, focusing on those with two gait analysis visits for supervised machine learning objectives.

**Figure 4** depicts the relationship between the ROC for two gait variables and the weeks elapsed between the injury and the initial gait analysis. This figure provides information for all fracture types and then specifically for Femur (Proximal, Diaphyseal, Distal) and Tibia (Proximal, Diaphyseal, Distal, Malleolar) fractures.

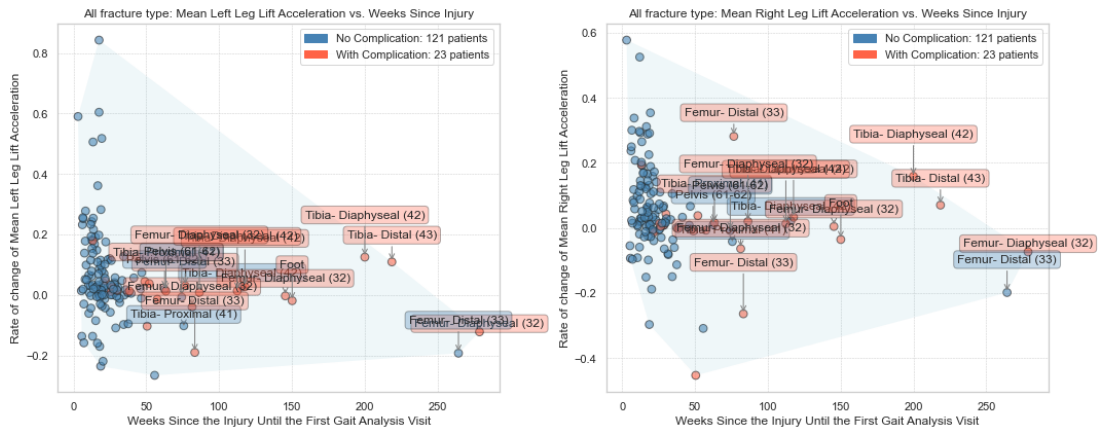

**Figure 4(a).** All fractures: Relationship between the ROC of two gait variables and the weeks elapsed post-injury.

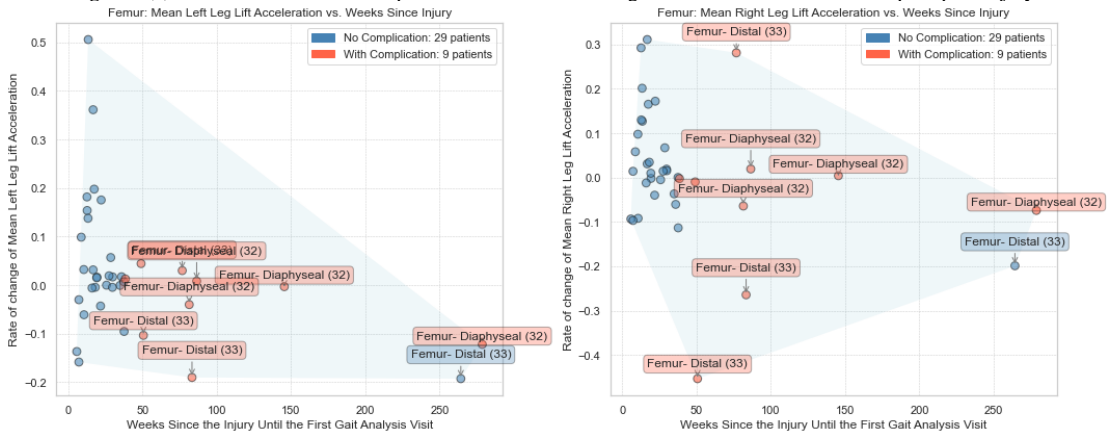

**Figure 4(b).** Femur fractures: Relationship between the ROC of two gait variables and the weeks elapsed post-injury.

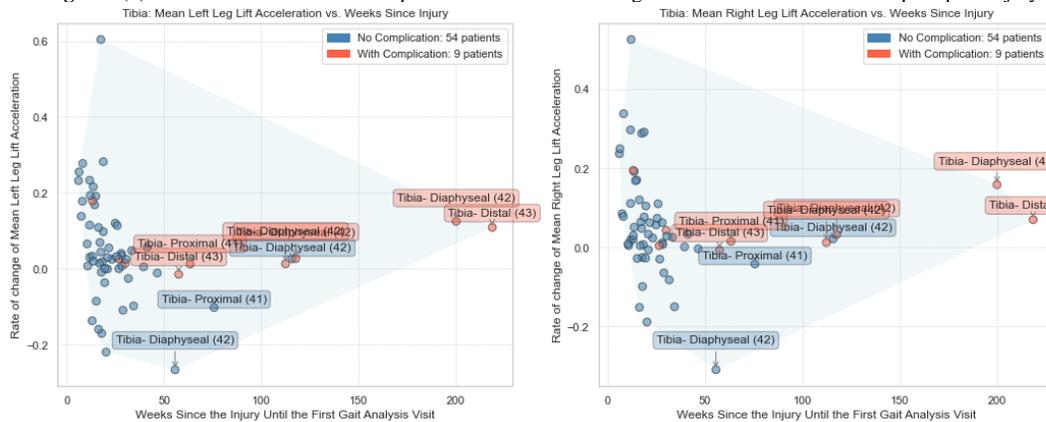

**Figure 4(c).** Tibia fractures: Relationship between the ROC of two gait variables and the weeks elapsed post-injury.

**Figure 4.** Relationship between the ROC of two gait variables and the weeks elapsed post-injury to the initial gait analysis, encompassing all fracture types and specifically highlighting Femur (Proximal, Diaphyseal, Distal) and Tibia (Proximal, Diaphyseal, Distal, Malleolar) fractures.

**Figure 5(a)** showcases ROCAUC curves, average of AUC, and accuracy scores for various machine learning models, including Logistic Regression (LR), Random Forest, XGBoost, LightGBM, and SVM, with 95% confidence intervals derived using a 1000-bootstrap method. Each curve uses hyperparameters optimized during the training-validation process. The figure contrasts model performances on training, validation, and test sets before applying SMOTE. After hyperparameter tuning, the selected hyperparameters for each model are as follows: LR is trained using L2-regularization (penalty). Random Forest has 3 trees with a maximum depth of 2. XGBoost has 5 trees with a maximum depth of 2. LightGBM has 3 trees with a maximum depth of 2. SVM uses a polynomial kernel with a degree of 3. **Figure 5(b)** presents the corresponding results after applying SMOTE to the training data. The XGBoost emerges as the best-fit model both before and after applying SMOTE. Before SMOTE's application, it yielded an average test AUC of 0.90 (95% CI: [0.79, 1.00]) and an average test accuracy of 86% (95% CI: [75%, 97%]). After applying SMOTE, the average AUC was 0.74 (95% CI: [0.69, 0.97]) and the average test accuracy was 79% (95% CI: [62%, 96%]). In **Figure 6**, we present the feature importance as determined by the logistic regression, SVM, and the XGBoost models before any SMOTE application.

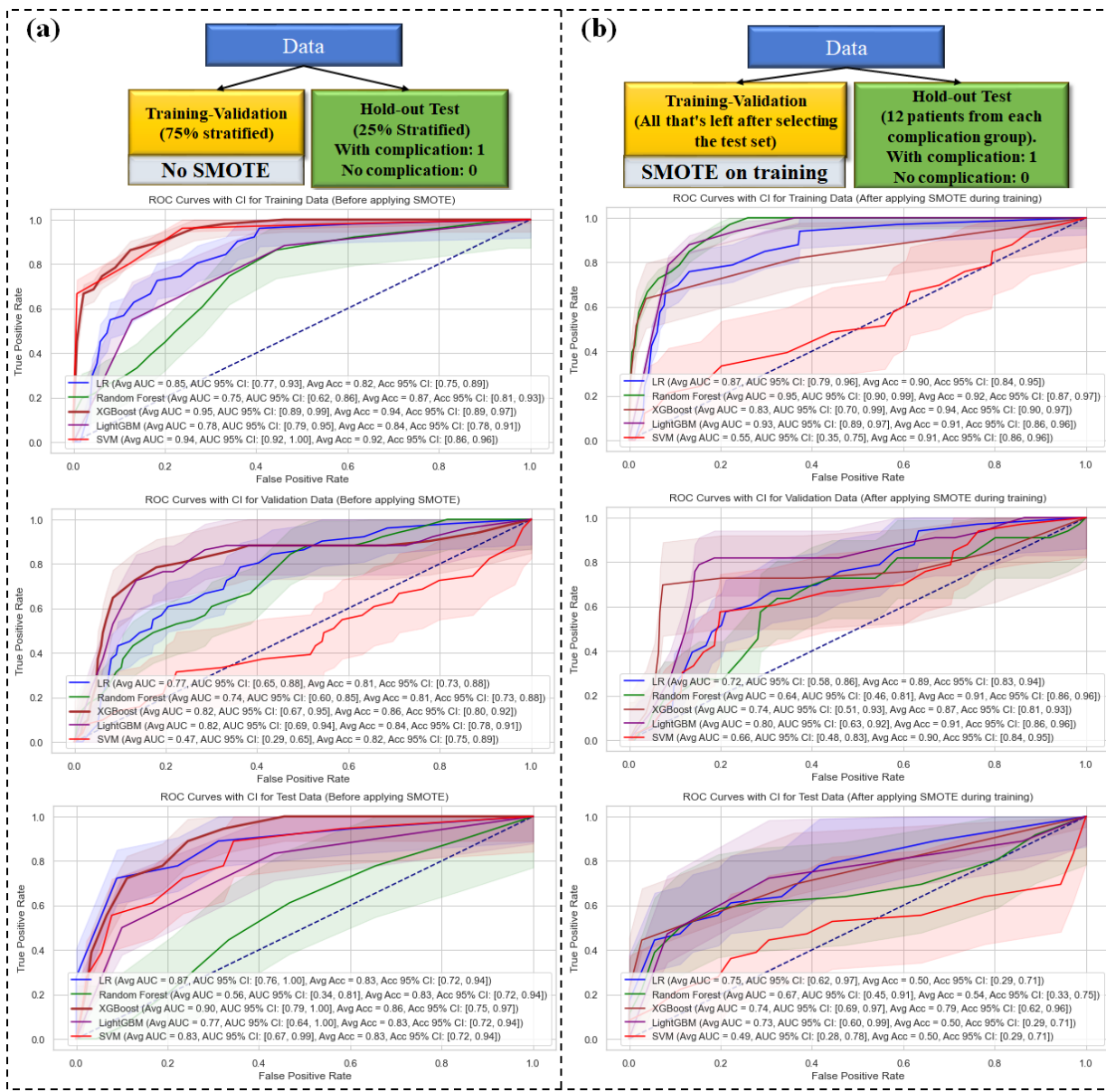

**Figure 5.** ROCAUC curves, AUC, and accuracy bars for machine learning models (before and after SMOTE application) with 95% confidence intervals derived from a 1000-bootstrap method. Model hyperparameters: LR: L2-regularization; Random Forest: 3 trees, max depth of 2; XGBoost: 5 trees, max depth of 2; LightGBM: 3 trees, max depth of 2; SVM: polynomial kernel, degree 3.

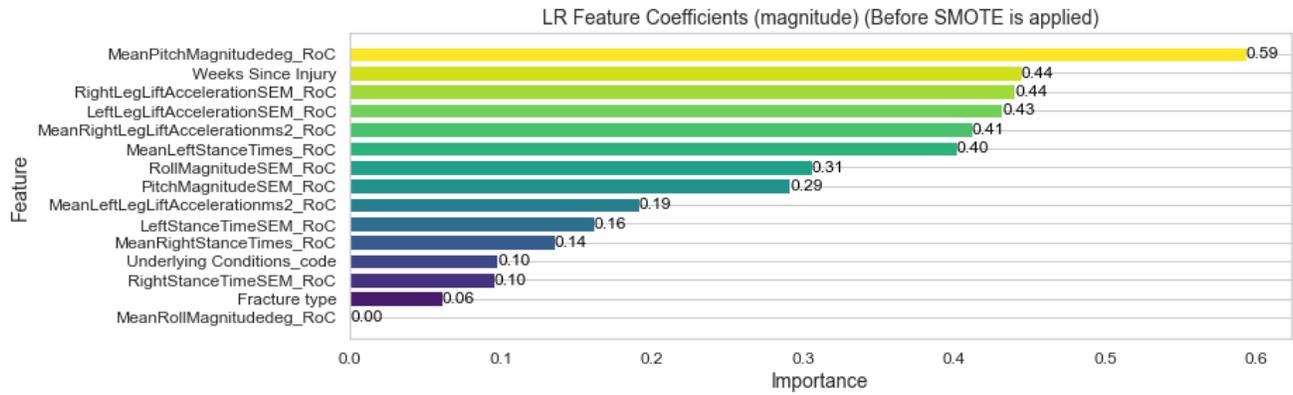

**Figure 6(a).** Depiction of feature importance as derived from **the logistic regression.**

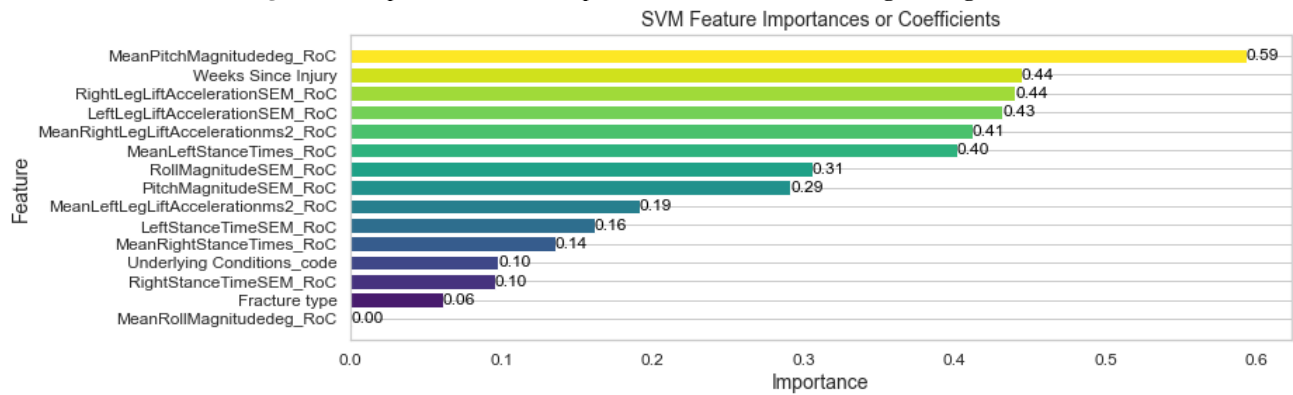

**Figure 6(b).** Depiction of feature importance as derived from the **SVM.**

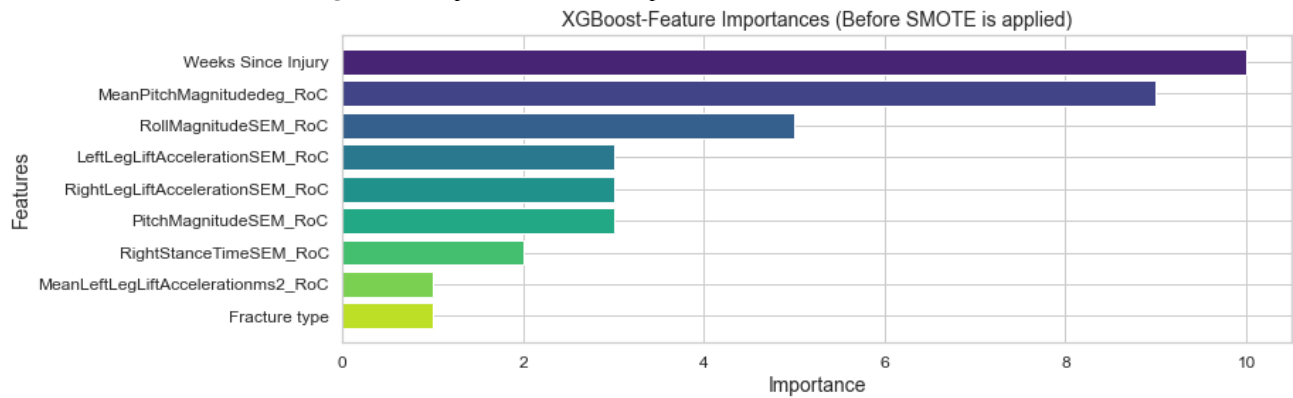

**Figure 6(a).** Depiction of feature importance as derived from the **XGBoost.**

**Figure 6.** Depiction of feature importance as derived from the logistic regression, SVM and the XGBoost models, prior to SMOTE application.

## 4 Discussion

The results from the Chi-squared test (as shown in **Figure 2**) reveal no significant association between underlying medical conditions and the main orthopedic complications, i.e., infection, malunion, or hardware irritation. However, a closer look at the data suggests that the proportion of patients with the underlying conditions who experience complications is slightly higher than those without any conditions (21.93% vs 15.49%). Conversely, the Chi-squared test did identify a significant connection between readmission rates and the presence of underlying medical conditions. Specifically, patients with these conditions were more likely to be readmitted, though not for common complications such as infection, malunion, or hardware irritation.

**Figure 4** delves deeper into the dynamics of post-injury recovery by examining the relationship between the number of weeks post-injury and the rate of change (ROC) in primary gait variables. A clear funnel-shaped pattern emerges, hinting at a brisk physiological response during the early stages after injury. The broader range of ROC values early on likely illustrates the body's immediate compensatory mechanisms. As the weeks progress, this range narrows, suggesting a stabilization or healing phase. The observed pattern underscores the vital role of gait analysis, especially in the early post-injury phase. By capturing a wide range of responses, it offers indispensable insights into the injury's repercussions on locomotion. Tracking these nuanced changes helps orthopedic professionals gauge the efficacy of early interventions and rehabilitation regimens, thereby enabling more tailored treatment decisions.

**Figure 5** underscores the superiority of XGBoost as the optimal modeling approach for the data, both before and after the application of SMOTE. The high accuracy and AUC averages from the hold-out test, derived after 1000 bootstraps, validate the efficacy of the IMU tool in the orthopedic domain. By processing IMU gait data from only two consecutive visits of patients who underwent gait analysis at varying time intervals and then converting these datasets into a unified feature vector using Algorithm 1, XGBoost demonstrated exceptional precision. Specifically, it could predict with remarkable accuracy whether a patient would experience complications such as infection, malunion, or hardware irritation. These findings hold significant implications for enhancing patient care and predictive outcomes in orthopedics.

Delving deeper into the nuances of the XGBoost model, **Figure 6** elucidates the paramount importance of the ROC gait variables in accurately predicting patients' complication statuses. Notably, the duration between the injury occurrence and the initial gait analysis stands out as the most critical predictor. Harmonizing the insights from **Figures 4** and **5**, a profound revelation emerges: Patients who proactively track their orthopedic recovery progress, leveraging advanced tools such as gait analysis, substantially reduce their risk of complications. This underscores the imperative role of timely and informed intervention in the realm of orthopedics, potentially revolutionizing patient outcomes. **Figure 6** displays the feature importance for non-tree-based models such as LR and SVM, and these findings align with the importance results from XGBoost.

The revelations from these results serve as a beacon for the orthopedic field, particularly in the context of lower extremity injuries. The early detection of physiological responses and the body's compensatory mechanisms illuminate a crucial window for orthopedic interventions. Recognizing the pivotal moments post-injury can redefine the trajectory of recovery, ensuring that interventions are timely, effective, and catered to the individual's unique physiological responses. The delicate balance of the body's early aggressive compensations followed by stabilization phases showcases the body's resilience and adaptability. Harnessing this knowledge can empower orthopedic professionals to finesse their therapeutic approaches, optimizing outcomes for patients.

As we integrate machine learning into orthopedic care, we open doors to unprecedented precision in patient management. The ability of models to predict potential complications with high accuracy signifies a monumental shift in how we approach orthopedic care. By identifying risks early on, medical professionals can take proactive steps, preventing complications before they manifest. This not only improves patient outcomes but also revolutionizes the patient experience, potentially reducing prolonged discomfort, additional surgeries, or intensive rehabilitation. The harmony between data-driven insights and clinical expertise paints a promising picture for the future of orthopedics. Embracing these advancements paves the way for a more informed, proactive, and patient-centric approach to care.

## 5      Conclusions

This study underscores the revolutionary potential of integrating machine learning techniques, specifically models like XGBoost, with gait analysis in the orthopedic landscape. By accurately predicting post-injury complications from consecutive gait datasets, we not only refine our diagnostic precision but also gain the ability to tailor interventions more responsively. The prominence of factors like the interval between injury onset and initial gait analysis signals a pressing need for early assessment, as discernible physiological compensations can significantly impact patient prognosis. As the medical community strives for enhanced patient care, the findings from this study position early gait analysis, augmented by machine learning, as a linchpin in orthopedic care. In essence, we are at the dawn of a new era where data-driven insights can substantially elevate the quality, precision, and personalization of orthopedic interventions, heralding a future where patient outcomes are not just improved but optimized.

## 6 Data Availability

The datasets used and analyzed during the current study available from the corresponding author on reasonable request.

## 7 Conflict of Interest

The authors declare that the research was conducted in the absence of any commercial or financial relationships that could be construed as a potential conflict of interest.

## 8 Author Contributions

**M.R.:** Data preparation, Formal analysis, Methodology, Software, Validation, Writing-original draft, Writing-review & editing. **R.B.S.:** Study Design, Writing-review & editing. **S.H.S.:** Study Design, Writing-review & editing. **M.A.K.:** Study Design, Writing-review & editing. **N.H.:** Data Collection. **M.N.G.:** Conceptualization, Formal analysis, Methodology, Validation, Project administration, Writing-original draft, Writing-review & editing.

## 9 Funding

This study was conducted with funding from the AO North America via the AO Strategy Fund.